\documentclass[conference]{IEEEtran}
\IEEEoverridecommandlockouts
\renewcommand\IEEEkeywordsname{Keywords}
\usepackage{fancyhdr}

\usepackage{cite}
\usepackage{amsmath,amssymb,amsfonts}
\usepackage{algorithmicx}
\usepackage{graphicx}
\usepackage{textcomp}
\usepackage{eso-pic}
\usepackage{xcolor}

\usepackage{tabularx}

\usepackage[ruled,vlined,noend]{algorithm2e}
\usepackage{array}

\usepackage{multirow}
\usepackage{algpseudocode}
\usepackage{flushend}
\newcommand\Mark[1]{\textsuperscript#1}

\fancypagestyle{firstpage}{
  \fancyhf{} % Clear all header and footer fields
  \fancyhead[L]{\small 2025 International Conference on Quantum Photonics, Artificial Intelligence, and Networking (QPAIN) \\ 31 July - 2 August 2025, Rangpur, Bangladesh}
  \fancyfoot[L]{\small 979-8-3315-9694-1/25/\$31.00 \copyright2025 IEEE} % Left-align the footer
}
\author{\IEEEauthorblockN{Md Gulzar Hussain\Mark{1},
Babe Sultana\Mark{2}, Md Rinku Ali\Mark{3}%, Choyon Das\Mark{4}, Md Sanzid Ahmed\Mark{5}
}

\IEEEauthorblockA{\Mark{1}School of Software, Nanjing University of Information Science and Technology, Nanjing, China\\
\Mark{1}$^,$\Mark{2}Department of Computer Science and Engineering, Green University of Bangladesh, Dhaka, Bangladesh\\ 
\Mark{3}School of Computer Science and Artificial Intelligence, Changzhou University, Changzhou, China\\
\Mark{1}gulzar.ace@gmail.com, \Mark{2}babecse@gmail.com, \Mark{3}rinkufast302550.ra@gmail.com%,\\ \Mark{4}choyondas08@gmail.com, \Mark{5}mdsanzidahmedsakib@outlook.com
}
}

\title{Hybrid Feature Combinations with CNN for Bangla Fake News Classification}
\begin{document}

\maketitle
%\thispagestyle{firstpage} % Apply the footer only on the first page

\begin{abstract}
Nowadays, people in Bangladesh frequently rely on the internet and social media for daily news instead of traditional newspapers. However, the spread of false Bangla news through these platforms poses risks and challenges to the credibility of authentic media. Although several studies have been conducted on detecting Bangla fake news, there is still significant room for improvement in this area. To assist people, this research explores the effectiveness of feature selection approaches in identifying appropriate features, such as semantic, statistical, and character-level features, or their combinations, on the BanFakeNews-2.0 dataset for detecting Bangla fake news using a CNN model. In this paper, key findings reveal that combining multiple features significantly improves recall and F1-scores compared to using individual features alone. The code for this research can be availed here, https://github.com/gulzar09/Bn\_FNews\_H.Feature.
\end{abstract}


\IEEEpeerreviewmaketitle

%------------------------------------------------------------------------------------------------>>>>>
\section{Introduction}\label{introduction}
News forms an essential aspect of social life and holds great importance in the daily routines of individuals. In the digital era, news flows rapidly and reaches people through various social media platforms and online news portals, a trend that is also prevalent in Bangladesh. This facility can also spread Bangla fake news, which can negatively affect people's daily lives. False news can harm people's daily lives. So, the detection of Bangla fake news has significant importance. However, detecting fake news is challenging due to issues of authenticity and intent to influence public opinion. In Bangla text, this is more complex due to its different writing structures and vocabulary. Though many works have already been done in English fake news detection, not many works have been done in the Bangla language \cite{hossain2021study}. Fake news detection from text is a subarea of text classification that requires high resources. In Bangla fake news, it requires more resources due to its high diversity and vocabulary. So, feature extraction from the texts has a high impact on this research work. A high number of features can influence the efficiency and performance of classification models as all features do not have the same significance and contribution to the model \cite{fayaz2022machine}. For this reason, removing less important features can increase the model's performance and reduce its complexity.

Recent studies used only single or a few feature extraction techniques like Count Vectorizer \cite{rasel2022bangla}, TF-IDF Vectorizer \cite{khatun2024bangla}\cite{hussain2020detection}, word2vec \cite{wahid2022bnnetxtreme}, GloVe \cite{wahid2022bnnetxtreme}\cite{hossain2021study}, fastText \cite{barua2025comparative} etc. but not their hybrids for Bangla fake news classification. Moreover, none of them compared the effectiveness of the feature extraction techniques in this task. Finding effective features can make the classification easier and more cost-effective. To fill this gap, in this study, we explore the effectiveness of some feature extraction approaches for Bangla fake news classification.\\

The structure of this paper is as follows: Section \ref{sec: related} presents an overview of the related background studies. In Section \ref{sec: method}, we describe the architecture of the proposed models in detail. Section \ref{sec: eva} evaluates the performance of these models using metrics such as accuracy, precision, recall, and F1-score. Lastly, Section \ref{sec: con} concludes the paper and outlines possible directions for future research.

%------------------------------------------------------------------------------------------------>>>>>
\section{Related Works}\label{sec: related}
Although numerous studies have been conducted on fake news detection, relatively few focus specifically on Bangla fake news. In this section, we provide a brief overview of some recent works related to Bangla fake news detection. Article \cite{rasel2022bangla} employed various machine learning approaches, including deep neural networks and transformer-based architectures, incorporating both Count Vectorizer and TF-IDF Vectorizer techniques for feature extraction, to detect fake news across both their own and external datasets, revealing better performance from deep learning methodologies. Another study \cite{rohman2023ibfnd} presented the Improved Bangla Fake News Dataset (IBFND) and evaluated Multinomial Naive Bayes, Linear Support Vector Classifier, XGBoost, and BERT with TF-IDF feature extraction, achieving an F1-score of 97\% on the dataset. The BNnetXtreme \cite{wahid2022bnnetxtreme} model enhanced classification accuracy by 1.1\%, integrating word embedding methods word2vec, GloVe, and fastText, for feature extraction in combination with a Bengali BERT model. By integrating TF-IDF, Bag of Words (BOW), N-Grams, and Word Embedding feature extraction with graph neural networks (GCN) and RoBERTa, the RoBERTa-GCN \cite{ahammad2024roberta} model achieved an accuracy of 98.60\% in Bangla fake news classification. Among various machine learning models evaluated for the same task, the Bangla-bert-base model combined with TF-IDF and Bag of Words feature extraction attained an accuracy of 84.19\%. The researchers in the article \cite{mondal2024breaking} employed label encoding to train a Gated Recurrent Unit (GRU) network for Bangla fake news detection, achieving an accuracy rate of 94\%. Bidirectional GRU with Tokenization and Padding techniques demonstrated exceptional performance \cite{roy2024enhancing}, achieving an accuracy of 99.16\% among various deep learning models for Bangla fake news detection. 

Some recent research has demonstrated exceptionally high performance in Bangla fake news classification through various advanced approaches, including the integration of the application of SMOTE with FastText and BERT \cite{barua2025comparative}, using Generative Adversarial Networks (GANs) and the FixMatch algorithm \cite{absar2025semi}, integrating tokenization and padding techniques, knowledge-based manipulation detection methods \cite{akther2025automatic}, and multimodal approaches that combine textual and visual information through DenseNet-169 and mBERT models \cite{faria2025multibanfakedetect}. Among the research discussed, none of them determined the effectiveness of feature extraction techniques that might have improved their performance, and we focus on addressing this gap in this research work.

\section{Proposed Method}\label{sec: method}
\begin{figure}[htp]
    \centering
   \includegraphics[width=8.5cm,height=7.0cm]{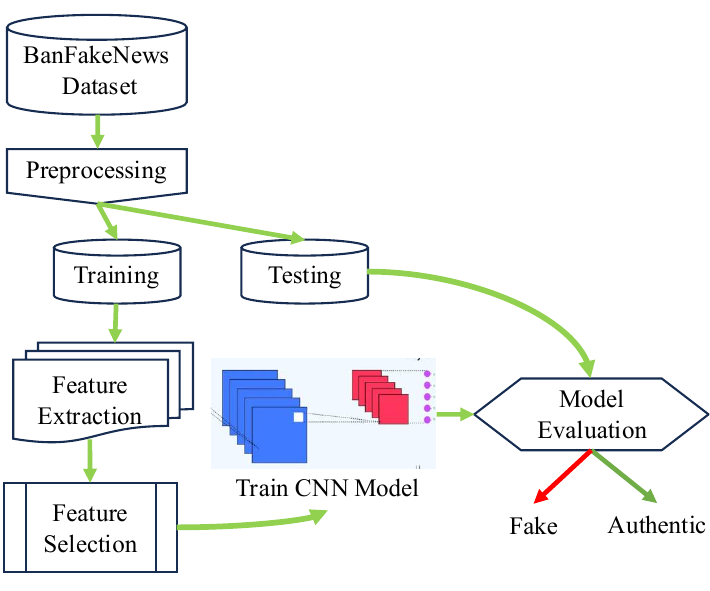}
    \caption{Proposed Research Flow Diagram.}
    \label{systemFlow}
\end{figure}

This research focuses on a crucial Bangla text classification task: Bangla fake news classification. The proposed architecture is illustrated in Fig \ref{systemFlow}. We have provided a detailed description of each step below.

\subsection{Dataset}
The dataset used in this research is the BanFakeNews-2.0 corpus \cite{shibu-etal-2025-scarcity}, a Bangla fake news dataset comprising 48,678 real news articles and 12,903 manually annotated fake news articles. The original dataset included separate training, validation, and test sets. For our experiments, we combined them into a single file and labeled real news as 1 and fake news as 0. After applying several preprocessing steps, we split the data into training (65\%), validation (15\%), and test (20\%) sets.

\subsection{Data Preprocessing}
Text preprocessing plays a crucial role in text classification, as irrelevant elements can negatively impact model performance. In this research, we have applied several preprocessing steps to clean the Bangla news texts. Specifically, we removed punctuation marks, Bangla stop words (common words), unnecessary special characters, and HTML tags. However, we retained digits, as they may carry important information in news content. After completing the preprocessing, we applied various feature extraction approaches to prepare the data for classification.

\subsection{Feature Extraction}
Machine Learning (ML) and Deep Learning (DL) models can only process numerical data, not raw textual input. Therefore, textual data must be converted into numerical representations using feature extraction techniques. These methods not only transform text into a suitable format but also help optimize the features for better performance in ML and DL models. We extracted features from the Bangla real and fake news data using the following six feature extraction methods.

\subsubsection{TF-IDF}
The TF-IDF is a popular statistical metric that indicates a word's significance to a document within a corpus or collection. The term frequency (TF) is calculated using the following formula in equation \ref{tf_equation}.

\begin{equation}
    tf(t, d) = \frac{In\hspace{0.1cm}document \hspace{0.1cm}d\hspace{0.1cm}count \hspace{0.1cm}of\hspace{0.1cm}term \hspace{0.1cm}t}{Total\hspace{0.1cm}number\hspace{0.1cm}of\hspace{0.1cm}terms\hspace{0.1cm}in\hspace{0.1cm}d}
    \label{tf_equation}
\end{equation}

The inverse document frequency (IDF) is calculated using the following formula in equation \ref{idf_equation}.

\begin{equation}
    idf(t) = \log\frac{1 + N}{1 + n_t} + 1
    \label{idf_equation}
\end{equation}
where, $N$ is total number of documents and $N_t$ is number of documents containing term $t$. Using the equations \ref{tf_equation} and \ref{idf_equation}, the final TF-IDF is calculated as in equation \ref{tf-idf_equation} \cite{hussain2020detection}.

\begin{equation}
    tf-idf(t, d) = tf (t, d) * idf(t)
    \label{tf-idf_equation}
\end{equation}

In this research, to extract TF-IDF features, we selected the top 5,000 most frequent tokens, considered both unigrams and bigrams, excluded words that appeared in two or fewer documents, and also removed words that appeared in more than 95\% of the documents.

\subsubsection{Word2Vec}
The neural embedding system Word2Vec learns to encode every word in a continuous vector space, with semantically related words grouped closer together. We used the CBOW (Continuous Bag of Words) architecture for the Word2Vec feature extraction method \cite{xia2023continuous}, setting the vector size of each word to 100. Additionally, all words that appeared at least once were included in the vocabulary. A context window of 5 words was used, training was performed using 4 parallel threads, and a random seed of 42 was set to ensure reproducibility.

\subsubsection{FastText}
Facebook AI Research (FAIR) created FastText as an enhancement to Word2Vec \cite{young2019review}.  FastText improves on Word2Vec's ability to learn embeddings by learning embeddings for subwords in addition to complete words. This enables FastText to create word vectors from subwords for uncommon or unseen words. In this research, we used the same configuration as applied in the Word2Vec-based feature extraction.

\subsubsection{N-Gram Features}
N-grams capture word sequences, helping to encode the structural and contextual information within text. In this research, we utilized CountVectorizer's n-gram extraction to transform input text into numerical feature vectors \cite{shibu-etal-2025-scarcity}. CountVectorizer constructs a document-term matrix, where each row represents a document and each column corresponds to an n-gram, thereby expanding the vocabulary based on observed n-grams. We extracted n-grams up to size 3, retained the top 1,000 most frequent n-grams, and excluded those that appeared in fewer than two documents.

\subsubsection{Character-Level TF-IDF Features}
Instead of extracting sequences of words, the character n-gram approach focuses on sequences of characters, giving more emphasis to informative subword patterns. In this research, we have computed character-level TF-IDF features using the same formula presented in equation \ref{tf-idf_equation}. We extracted character n-grams ranging from 2 to 4 characters in length, retained the top 1,000 most frequent n-grams, and excluded those that appeared in two or fewer documents.

\subsubsection{Statistical Text Features}
Statistical features are numerical descriptors that are manually constructed and describe various characteristics of the text.  They rely on the form, structure, and composition of the text rather than the semantic content of words, in contrast to deep or distributional characteristics. In this research, the following statistical features were calculated for each news article: character count, word count, sentence count, average word length, punctuation count, digit count, average characters per word, and punctuation ratio.

\subsection{Feature Selection}
Generally, it is not a good idea to use all the statistical features extracted in any research, as they may not all hold the same level of significance or value. Some features might be more informative and contribute more effectively to fake news classification and overall model performance, while others may be less valuable. Moreover, selecting the right set of features can help reduce model overfitting and improve precision \cite{sverdrup2021feature}. For this reason, we applied five feature selection approaches: Statistical, Ensemble-Based Importance, Correlation-Based Selection, Forward Selection, and Comprehensive Combination Testing.

\subsubsection{Statistical Feature Selection}
We employed three statistical feature selection methods—ANOVA F-test, Chi-Square test, and Mutual Information—to identify the most informative features from the combined feature sets. The ANOVA F-test determines whether the mean values of a feature differ significantly across target classes, the Chi-Square test measures the dependency between features and target labels, and Mutual Information assesses the amount of information a feature provides about the target by calculating information gain.

\subsubsection{Ensemble-Based Feature Importance}
Ensemble models such as Random Forest and Extra Trees are highly effective for estimating feature importance. In this research, we used Random Forest to determine the importance scores of features concerning the target labels. The model calculates these scores based on each feature's contribution to decision-making across all trees in the forest, and then computes the average importance for each feature.

\subsubsection{Recursive Feature Elimination (RFE)}
We employed a wrapper-based Recursive Feature Elimination (RFE) approach to identify the most important features for Bangla fake news classification. In our method, RFE uses logistic regression as the base estimator to rank feature importance, progressively eliminating the least important features. This process continues iteratively until the desired number of features—six in our case—is selected.

\subsubsection{Correlation-Based Feature Selection}
In order to determine how strongly each set of characteristics correlates with the target variable, we employ a filter-based correlation-based feature selection method. It achieves this by evaluating the linear connection between variables using the Pearson correlation coefficient, a commonly used statistical metric. Equation \ref{pearson_equation} is used in this study to calculate the Pearson Correlation Coefficient (r), and the absolute value of $r$ is used to evaluate the strength of the relationship, independent of its direction.

\begin{equation}
    r = \frac{\sum_{i=1}^{n}(x_i-\bar{x})(y_i-\bar{y})}{\sqrt{\overline{\sum_{i=1}^{n}(x_i-\bar{x})^2}}\sqrt{\overline{\sum_{i=1}^{n}(y_i-\bar{y})^2}}}
    \label{pearson_equation}
\end{equation}
where $x_i$ is the value of the feature and $y_i$ is the corresponding target label for the $i^{th}$ sample, $\bar{x}$ \& $\bar{y}$ are the mean of the feature \& target values, and n is the total number of samples.

\subsubsection{Forward Feature Selection}
We used a wrapper-based forward selection technique in Forward Feature Selection, which uses model performance on a validation set to gradually choose the most informative feature sets. We used logistic regression as the base model and evaluated performance using the F1-score on the validation set. The forward selection process begins with an empty set of features and adds one feature at a time, selecting the one that leads to the greatest improvement in validation performance. The feature contributing the highest F1-score at each step is added to the final selected feature set.

\subsubsection{Comprehensive Feature Testing}
We conducted a brute-force feature combination test by evaluating all possible subsets of the six extracted features to identify the most effective combination. Logistic Regression was used as the classifier, and the performance was assessed using the F1-score for fake news on the validation set. The procedure for this approach is outlined in Algorithm \ref{CFCT_algorithm}.

\begin{algorithm}[ht]
\caption{Comprehensive Feature Combination Testing}
\SetAlgoNoLine
\begin{algorithmic}[1]
\Require Training features $\mathcal{F}_{train}$, Validation features $\mathcal{F}_{val}$, Labels $y_{train}$, $y_{val}$, Minimum size $k$
\Ensure Top feature combinations ranked by fake news F1-score

\State Initialize result list $\mathcal{R} \gets []$
\State Extract feature names $\mathcal{F} \gets$ keys of $\mathcal{F}_{train}$

\hspace{-2.5em}\For{$r = k$ to $|\mathcal{F}|$}{
    \hspace{-1.5em}\For{Feature combinations $C$ of size $r$}{
        \State $X_{train} \gets$ concatenate features in $C$ from $\mathcal{F}_{train}$
        \State $X_{val} \gets$ concatenate features in $C$ from $\mathcal{F}_{val}$
        \State Train LR on $X_{train}$, predict $\hat{y}$ on $X_{val}$
        \State Compute accuracy, F1$_\text{fake}$, F1$_\text{real}$, F1$_\text{macro}$
        \State Append $(C, \text{scores})$ to $\mathcal{R}$
    }
}
\vspace{0.3em}
\State Sort $\mathcal{R}$ by F1$_\text{fake}$ (descending)
\State \Return Top combinations
\end{algorithmic}
\label{CFCT_algorithm}
\end{algorithm}

\subsection{CNN Architecture}
The suggested CNN model has a multi-branch design in which several convolutional pipelines handle different kinds of input features separately.  Each branch compresses high-dimensional feature representations into compact vectors and extracts local patterns using global max pooling and 1D convolutions.  Concatenating these vectors and passing them through fully connected layers allows for the efficient binary classification of authentic and fake Bangla news by learning complicated relationships. Table \ref{hperP_CNN} shows the hyperparameters used in this proposed CNN architecture.

\begin{table}[htp]
    \centering
    \caption{Hyper Parameter Settings of Proposed CNN Model.}
    \begin{tabular}{>{\raggedright\arraybackslash}m{3cm} >{\raggedright\arraybackslash}m{4.7cm}}
    \hline
       \textbf{Hyperparameter}  &  \textbf{Value} \\ \hline
       Number of Epochs & 50 \\ \hline
       Loss Function & Binary Crossentropy \\ \hline
       Evaluation Metric & Accuracy, Precision, Recall \\ \hline
       Optimizer  &  Adam \\ \hline
       Initial Learning Rate & 0.0001 \\ \hline
       Batch Size & 16 \\ \hline
       Early Stopping Patience	& 5 epochs ($monitor = val\_loss$) \\ \hline
       Learning Rate Scheduler	& ReduceLROnPlateau ($factor=0.5, patience=3$) \\ \hline
       Dropout & 0.3 \\ \hline
       Fully Connected Layer 1	& 256 units + Dropout(0.5) \\ \hline
       Fully Connected Layer 2	& 128 units + Dropout(0.3) \\ \hline
       Fully Connected Layer 3 & 64 units \\ \hline
       Output Layer & 1 unit + Sigmoid \\ \hline
        
    \end{tabular}
    \label{hperP_CNN}
\end{table}

\section{Experimental Results}\label{sec: eva}
This section first analyzes the outcomes of the feature selection methods. Based on their performance, the most effective feature sets are selected. These selected sets are then used to evaluate the proposed CNN model on the BanFakeNews-2.0 dataset.

\subsection{Results of Feature Selection}
Table \ref{res_feature_selection} presents the evaluation results of the Statistical, Ensemble-Based, Recursive Feature Elimination, and Correlation-Based feature selection techniques on the BanFakeNews-2 dataset.

\begin{table}[ht]
\caption{Results of the Statistical, Ensemble-Based, Recursive Feature Elimination, and Correlation-Based feature selection techniques}
\centering
\begin{tabular}{>{\raggedright\arraybackslash}m{0.9cm}>{\raggedright\arraybackslash}m{1.2cm}>{\raggedright\arraybackslash}m{1.0cm}>{\raggedright\arraybackslash}m{1.2cm}>{\raggedright\arraybackslash}m{1.3cm}>{\raggedright\arraybackslash}m{0.8cm}}
\hline 
\textbf{Feature}  & \textbf{F-CLASSIF} & \textbf{RF Importance} & \textbf{Correlation Score} & \textbf{Forward Selection} & \textbf{Overall Rank} \\ \hline
FastText & \textbf{2 selected}           & \textbf{0.0026} & \textbf{0.1387}           & \textbf{Step 2 (F1 to 0.7395)}    & \textbf{1st}                  \\ \hline
Word2Vec & 1 selected           & \textbf{0.0026} & 0.1331                     & Step 4 (F1 to 0.7472)    & 2nd                   \\ \hline
Char lvl Tf-Idf     & 1 selected           & 0.0008        & 0.0526                     & Step 3 (F1 to 0.7436)    & 3rd                   \\ \hline
TF-IDF   & 0 selected           & 0.0001        & 0.0277                     & Step 1 (F1 to 0.7024)    & 4th \\ \hline
N-gram   & 0 selected           & 0.0001        & 0.0304                     & Not selected              & 5th            \\ \hline
Statistical    & 0 selected           & 0.0011        & NaN                        & Not selected              & 6th            \\ \hline
\end{tabular}
\label{res_feature_selection}
\end{table}

From table \ref{res_feature_selection}, it is evident that FastText is the most influential feature among the six evaluated. Word2Vec and character-level TF-IDF follow as the second and third best-performing features, respectively, while statistical features show the weakest performance. Across all feature evaluation methods, FastText and Word2Vec consistently achieved closely matched scores. On the other hand, table \ref{res_comprehensiveTop10} presents the top 10 feature combinations based on the evaluation results of the comprehensive feature testing performed on all six features using the dataset employed in this research.

\begin{table}[ht]
\caption{Top 10 Feature Combinations of Comprehensive Feature Testing by Fake News F1-score.}
\centering
\begin{tabular}{>{\raggedright\arraybackslash}m{5.6cm}>{\raggedright\arraybackslash}m{1.2cm}}
\hline 
\textbf{Feature Combination}  & \textbf{F1-score} \\ \hline
tfidf, fasttext, ngram, char, stats & 0.748 \\ \hline
tfidf, word2vec, fasttext, char & 0.747 \\ \hline
tfidf, word2vec, char & 0.747 \\ \hline
tfidf, word2vec, fasttext, char, stats & 0.747 \\ \hline
tfidf, fasttext, ngram, char & 0.747 \\ \hline
tfidf, word2vec, char, stats & 0.746 \\ \hline
tfidf, word2vec, fasttext, ngram, char, stats & 0.745 \\ \hline
tfidf, word2vec, fasttext, ngram, char & 0.744 \\ \hline
tfidf, word2vec, ngram, char & 0.744 \\ \hline
tfidf, fasttext, char & 0.744 \\ \hline

\end{tabular}
\label{res_comprehensiveTop10}
\end{table}

Based on the results from Tables \ref{res_feature_selection} and \ref{res_comprehensiveTop10}, we selected the top five feature combinations from Table \ref{res_comprehensiveTop10}, along with the top three individual features from Table \ref{res_feature_selection} and the pair of top two features of Table \ref{res_feature_selection} for Bangla fake news classification using the proposed CNN model.

\subsection{Comparison of the Proposed CNN with Selected Features}
To assess the classification performance of the CNN model using the selected features on the BanFakeNews-2.0 dataset, we employed common evaluation metrics: accuracy, precision, recall, and F1-score. Table \ref{res_comparisonCNN} presents the comparison results of the CNN classifier for the selected feature sets.

\begin{table}[ht]
\caption{Result Comparison of CNN for Selected Feature Combinations.}
\centering
\begin{tabular}{>{\raggedright\arraybackslash}m{2.8cm}>{\raggedright\arraybackslash}m{1.0cm}>{\raggedright\arraybackslash}m{1.0cm}>{\raggedright\arraybackslash}m{0.8cm}>{\raggedright\arraybackslash}m{1.1cm}}
\hline 
\textbf{Feature Combination}  & \textbf{Accuracy} & \textbf{Precision} & \textbf{Recall} & \textbf{F1-score} \\ \hline
fasttext & 0.89 & 0.87 & 0.77 & 0.80 \\ \hline
word2vec & 0.89 & 0.88 & 0.76 & 0.80 \\ \hline
char & 0.84	& 0.80 & 0.65 & 0.68 \\ \hline
word2vec, fasttext & 0.89 & 0.87 & 0.77 & 0.81 \\ \hline
tfidf, word2vec, char & 0.89 & 0.87	& 0.78 & 0.81 \\ \hline
tfidf, word2vec, fasttext, char & 0.89 & 0.89 & 0.78 & 0.81 \\ \hline
tfidf, fasttext, ngram, char & 0.89	& 0.88 & 0.78	& 0.81 \\ \hline
tfidf, fasttext, ngram, char, stats & 0.90	& 0.89 & \textbf{0.79} & 0.82 \\ \hline
tfidf, word2vec, fasttext, char, stats & \textbf{0.91} & \textbf{0.90} & \textbf{0.79} & \textbf{0.83} \\ \hline

\end{tabular}
\label{res_comparisonCNN}
\end{table}

\begin{figure*}[ht]
    \centering
\includegraphics[width=15.5cm,height=11.0cm]{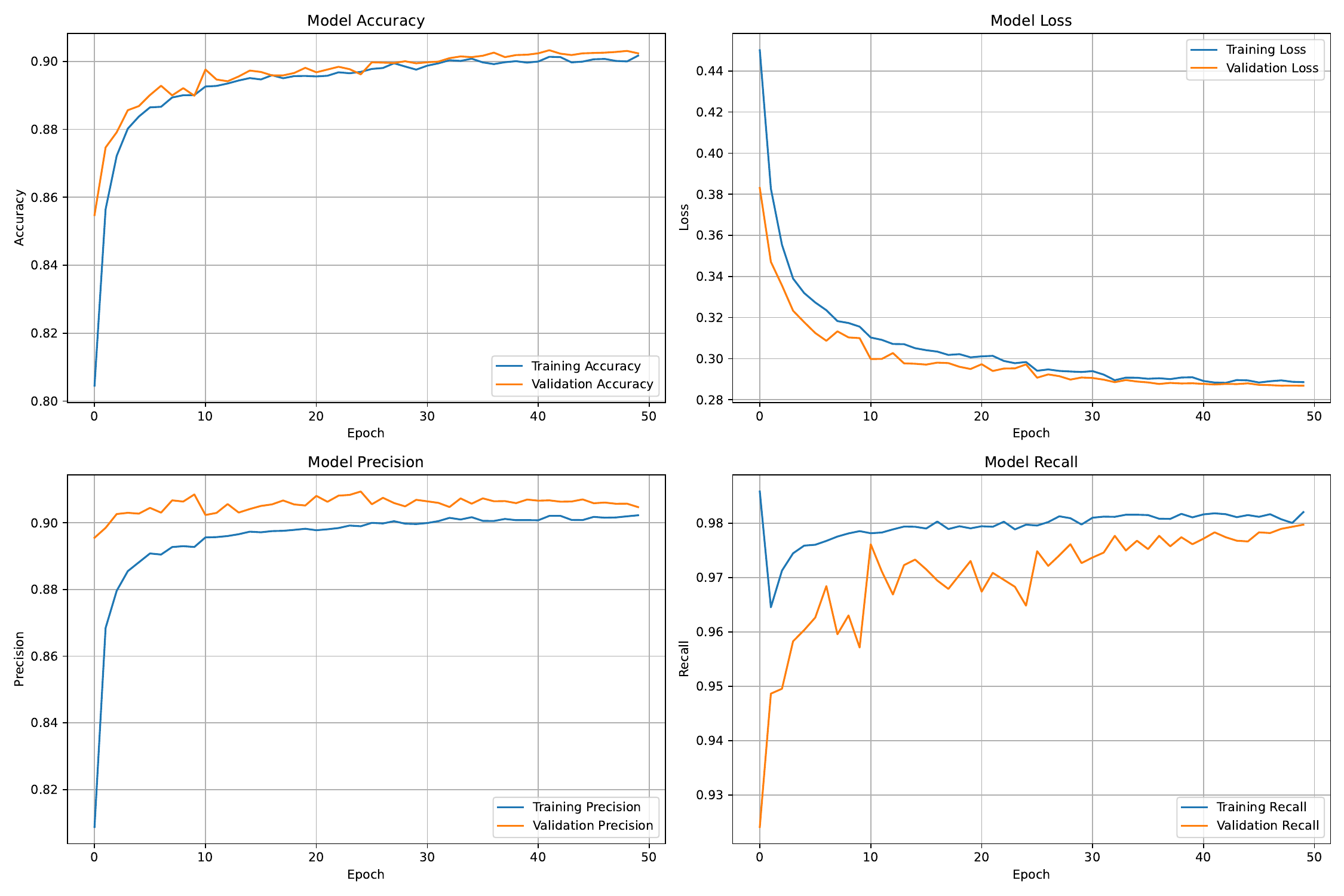}
    \caption{Training and Validation History for TF-IDF, Word2Vec, FastText, Character-level TF-IDF, and statistical characteristics Combination.}
    \label{trainHistBest}
\end{figure*}

Observing Table \ref{res_comparisonCNN}, it is evident that FastText and Word2Vec individually performed similarly, and their combination slightly improved the performance, achieving an F1-score of 0.81. However, incorporating additional features further enhances the results, particularly in terms of F1-score and recall, which are critical for fake news detection. The combination of TF-IDF, Word2Vec, FastText, Character-level TF-IDF, and Statistical features achieved the highest scores across all metrics: accuracy (0.91), precision (0.90), recall (0.79), and F1-score (0.83). These findings indicate that integrating diverse feature types significantly boosts the model’s effectiveness in classifying Bangla fake news, with both semantic and statistical features contributing in a complementary way. The results become even clearer when examining Table \ref{res_comparisonFN}, which presents the precision, recall, and F1-score specifically for the fake news class using different combinations of features.

\begin{table}[ht]
\caption{Result Comparison of CNN for Selected Feature Combinations Only for Fake News Class.}
\centering
\begin{tabular}{>{\raggedright\arraybackslash}m{4.0cm}>{\raggedright\arraybackslash}m{1.0cm}>{\raggedright\arraybackslash}m{1.0cm}>{\raggedright\arraybackslash}m{1.1cm}}
\hline 
\textbf{Feature Combination}  & \textbf{Precision} & \textbf{Recall} & \textbf{F1-score} \\ \hline
fasttext & 0.84 & 0.56 & 0.67 \\ \hline
word2vec & 0.86 & 0.55 & 0.67 \\ \hline
char & 0.75 & 0.32 & 0.45 \\ \hline
word2vec, fasttext & 0.84 & 0.57 & 0.68 \\ \hline
tfidf, word2vec, char & 0.85	& 0.59 & 0.69 \\ \hline
tfidf, word2vec, fasttext, char & 0.88 & 0.57 & 0.69 \\ \hline
tfidf, fasttext, ngram, char	& 0.85 & 0.58	& 0.69 \\ \hline
tfidf, fasttext, ngram, char, stats & 0.86 & \textbf{0.60} & 0.70 \\ \hline
tfidf, word2vec, fasttext, char, stats & \textbf{0.90} & \textbf{0.61} & \textbf{0.73} \\ \hline

\end{tabular}
\label{res_comparisonFN}
\end{table}

The findings from Table \ref{res_comparisonFN} clearly show that relying only on particular features, such as FastText, Word2Vec, or Character-level TF-IDF, leads to comparatively lower F1-scores and recall. For instance, Word2Vec by itself produced an F1-score of 0.67 with an accuracy of 0.86 and a poor recall of 0.55. On the other hand, performance is continuously improved by combining features. Integrating TF-IDF, Word2Vec, FastText, Character-level TF-IDF, and statistical characteristics achieved the most significant results, with the highest accuracy (0.90), recall (0.61), and F1-score (0.73). Fig. \ref{trainHistBest} illustrates the training and validation accuracy, loss, precision, and recall across epochs during the training of the CNN using this feature combination. The curves indicate stable learning behavior, with no signs of overfitting throughout the training process.

The results in Tables \ref{res_comparisonCNN} and \ref{res_comparisonFN} indicate that individual features such as FastText and Word2Vec perform reasonably well; however, their recall for fake news remains limited, often missing a significant portion of counterfeit instances. In contrast, combining multiple features—specifically TF-IDF, Word2Vec, FastText, character-level TF-IDF, and statistical features—leads to substantial improvements. The overall F1-score increases to 0.83, and the recall for fake news improves from approximately 0.55–0.56 to 0.61. These findings suggest that incorporating a diverse set of statistical and semantic features enhances the model’s capability to detect fake news with higher accuracy and robustness.

\section{Conclusion}\label{sec: con}
In this research, we investigated the effectiveness of several feature extraction techniques and their hybrid combinations, utilizing various feature selection approaches. We also evaluated the performance of the selected features and their combinations using a CNN-based classifier. We found that individual features like FastText and Word2Vec performed well; however, integrating multiple features—such as TF-IDF, Word2Vec, FastText, character-level TF-IDF, and statistical features—improved both the recall and F1-score for overall results and the fake news class specifically. However, this study used only CNN for classification, while other popular machine learning and deep learning methods could be explored as classifiers in future work. Additionally, more feature extraction techniques and feature selection approaches can be investigated in the future.

%\subsection{Future Work}

%------------------------------------------------------------------------------------------------>>>>>
\bibliographystyle{ieeetran}
\bibliography{references}

\end{document}